\documentclass[conference]{IEEEtran}
\IEEEoverridecommandlockouts
\usepackage{cite}
\usepackage{amsmath,amssymb,amsfonts}
\usepackage{algorithmic}
\usepackage{graphicx}
\usepackage{textcomp}
\usepackage{threeparttable}
\usepackage{xcolor}
\usepackage{multirow}
\usepackage{booktabs}

\makeatletter
\newcommand{\linebreakand}{%
  \end{@IEEEauthorhalign}
  \hfill\mbox{}\par
  \mbox{}\hfill\begin{@IEEEauthorhalign}
}
\makeatother

\def\BibTeX{{\rm B\kern-.05em{\sc i\kern-.025em b}\kern-.08em
    T\kern-.1667em\lower.7ex\hbox{E}\kern-.125emX}}
\begin{document}

\title{BatteryAgent: Synergizing Physics-Informed Interpretation with LLM Reasoning for Intelligent Battery Fault Diagnosis\\

\thanks{This work is supported by the National Key Research and Development Program of China (2025YFE0207400), the Tsinghua-Toyota Joint Research Fund, the National Natural Science Foundation of China (62273197 and 62403276), and the Beijing Natural Science Foundation (L233027). \\*Corresponding author: Benben Jiang.}
}

\author{
\IEEEauthorblockN{Songqi Zhou}
\IEEEauthorblockA{Department of Automation\\
Tsinghua University\\
Beijing, China\\
zhousongqi@mail.tsinghua.edu.cn}
\and
\IEEEauthorblockN{Ruixue Liu}
\IEEEauthorblockA{Department of Automation\\
Tsinghua University\\
Beijing, China\\
liuruixue@mail.tsinghua.edu.cn}
\and
\IEEEauthorblockN{Boman Su}
\IEEEauthorblockA{Department of Automation\\
Tsinghua University\\
Beijing, China\\
1030super@tsinghua.edu.cn}
\linebreakand
\IEEEauthorblockN{Jiazhou Wang}
\IEEEauthorblockA{Department of Automation\\
Tsinghua University\\
Beijing, China\\
wangjiazhou@mail.tsinghua.edu.cn}
\and
\IEEEauthorblockN{Yixing Wang}
\IEEEauthorblockA{Department of Automation\\
Tsinghua University\\
Beijing, China\\
yx-wang21@mails.tsinghua.edu.cn}
\and
\IEEEauthorblockN{Benben Jiang*}
\IEEEauthorblockA{Department of Automation\\
Tsinghua University\\
Beijing, China\\
bbjiang@tsinghua.edu.cn}
}
\maketitle

\begin{abstract}
Fault diagnosis of lithium-ion batteries is critical for system safety. While existing deep learning methods exhibit superior detection accuracy, their ``black-box'' nature hinders interpretability. Furthermore, restricted by binary classification paradigms, they struggle to provide root cause analysis and maintenance recommendations. To address these limitations, this paper proposes \textbf{BatteryAgent}, a hierarchical framework that integrates physical knowledge features with the reasoning capabilities of Large Language Models (LLMs). The framework comprises three core modules: (1) A \textbf{Physical Perception Layer} that utilizes 10 mechanism-based features derived from electrochemical principles, balancing dimensionality reduction with physical fidelity; (2) A \textbf{Detection and Attribution Layer} that employs Gradient Boosting Decision Trees and SHAP to quantify feature contributions; and (3) A \textbf{Reasoning and Diagnosis Layer} that leverages an LLM as the agent core. This layer constructs a ``numerical-semantic'' bridge, combining SHAP attributions with a mechanism knowledge base to generate comprehensive reports containing fault types, root cause analysis, and maintenance suggestions. Experimental results demonstrate that BatteryAgent effectively corrects misclassifications on hard boundary samples, achieving an AUROC of 0.986, which significantly outperforms current state-of-the-art methods. Moreover, the framework extends traditional binary detection to multi-type interpretable diagnosis, offering a new paradigm shift from ``passive detection'' to ``intelligent diagnosis'' for battery safety management.
\end{abstract}

\begin{IEEEkeywords}
Battery Fault Diagnosis, Large Language Models, Interpretability, AI Agent
\end{IEEEkeywords}

\begin{figure*}[ht]
    \centering
    \includegraphics[width=0.9\textwidth]{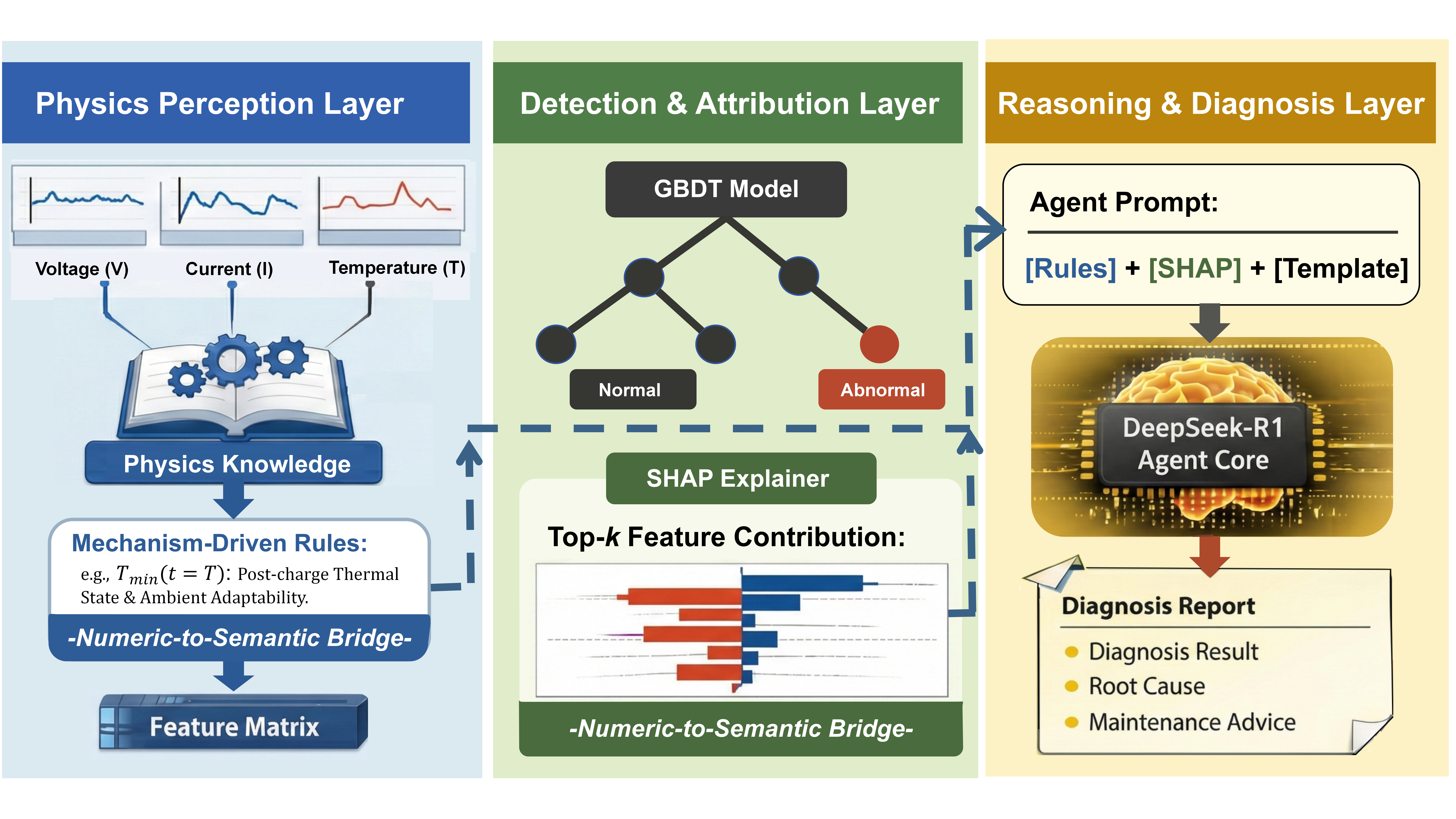}
    \caption{The BatteryAgent framework architecture. The framework comprises three layers: (1) Physics Perception Layer for mechanism-driven feature extraction, (2) Detection \& Attribution Layer using GBDT and SHAP for fault classification and feature contribution analysis, and (3) Reasoning \& Diagnosis Layer where an LLM agent generates diagnostic reports with root cause analysis and maintenance recommendations via the Numeric-to-Semantic Bridge.} \label{fig:framework}
\end{figure*}

\section{Introduction}

Lithium-ion batteries (LIBs) constitute the predominant energy storage technology for electric vehicles (EVs) owing to their superior energy density, extended cycle life, and broad operational temperature range \cite{b2}. The global EV market has experienced substantial growth, with sales surpassing 10 million units in 2022 and projected CO\textsubscript{2} reductions of approximately 700 million tons by 2030 \cite{b3}. This expansion is propelled by stringent emission regulations and substantial investments in battery manufacturing infrastructure \cite{b4}. However, higher energy density introduces commensurate safety risks: thermal runaway propagates rapidly and proves difficult to suppress \cite{b3}. Localized faults—including overcharge, over-discharge, internal short circuits, and anomalous degradation—can precipitate catastrophic failures manifesting as fires or explosions \cite{b2}. Given that high-power applications necessitate coordinated operation of numerous cells, single-cell malfunctions may compromise entire pack integrity. Consequently, battery management systems must continuously monitor voltage, current, and temperature to provide early fault warnings, yet accurate diagnosis under stochastic operating conditions remains a formidable challenge \cite{b2}.

Fault diagnosis methodologies bifurcate into physics-based and data-driven approaches~\cite{b5,b6}. Physics-based models, including electrochemical models and equivalent circuit models, offer mechanistic interpretability but impose substantial computational demands and parameterization requirements. Electrochemical models encounter convergence difficulties and often fail to capture complex real-world phenomena, while equivalent circuit models cannot represent internal electrochemical states such as side-reaction overpotentials \cite{b5}. Furthermore, model parameters exhibit dependencies on temperature and operating conditions, necessitating extensive experiments for online identification \cite{b8}. Data-driven methods provide scalable alternatives but require extensive high-quality data and function as opaque predictors \cite{b10}. The absence of physical constraints may yield predictions violating electrochemical principles or exhibiting poor generalization \cite{b12}. Moreover, existing approaches predominantly adopt binary classification paradigms, generating alarms without elucidating causative mechanisms or recommending maintenance actions \cite{b13,b14}.

Recent advances in large language models (LLMs) have demonstrated remarkable capabilities in natural language comprehension and complex reasoning~\cite{b15}. Through prompt engineering and chain-of-thought techniques, LLMs exhibit emergent abilities in systematic diagnostic reasoning, particularly in knowledge-intensive domains requiring multi-step inference and causal analysis. These capabilities suggest potential for bridging quantitative analytical outputs and human-interpretable diagnostic reports. However, LLMs exhibit well-documented limitations when processing numerical time-series data: empirical studies reveal persistent difficulties in multi-step numerical reasoning, statistical computations within specified ranges, and temporal pattern recognition~\cite{b16}. Direct ingestion of raw sensor sequences would incur prohibitive token costs while risking hallucinated or unreliable outputs. Consequently, establishing an effective interface between numerical measurements and semantic reasoning constitutes a critical prerequisite for leveraging LLMs in battery fault diagnostics.

To address these challenges, we propose \textbf{BatteryAgent}, a hierarchical framework integrating mechanism-driven feature engineering, interpretable machine learning, and LLM reasoning. The \textit{Physics Perception Layer} distills sensor streams into ten electrochemically-grounded features, reducing dimensionality while preserving physical fidelity. The \textit{Detection and Attribution Layer} employs Gradient Boosting Decision Tree (GBDT) classification with SHapley Additive exPlanations (SHAP)-based attribution for transparent feature contribution analysis. The \textit{Reasoning and Diagnosis Layer} utilizes an LLM agent that, via a numeric-to-semantic bridge, performs fault determination, causal analysis, and maintenance recommendation. An active refinement mechanism enables re-evaluation of low-confidence cases.

Our contributions are threefold: (i) BatteryAgent pioneers closed-loop collaboration among mechanistic features, interpretable ML, and LLM reasoning, enabling a shift from passive detection to active diagnosis; (ii) Experiments demonstrate an AUROC of 0.986, surpassing state-of-the-art baselines; (iii) The numeric-to-semantic bridge extends binary detection to multi-fault diagnosis with root-cause analysis and maintenance recommendations.

\section{Proposed Methodology}

\subsection{Framework Overview}

BatteryAgent is a hierarchical diagnostic framework that synergizes physics-informed feature engineering with LLM reasoning capabilities. As illustrated in Fig.~\ref{fig:framework}, the framework comprises three tightly coupled layers: \textit{Physics Perception Layer}, \textit{Detection \& Attribution Layer}, and \textit{Reasoning \& Diagnosis Layer}. The core innovation lies in establishing a ``Numeric-to-Semantic Bridge'' that transforms quantitative sensor measurements into interpretable diagnostic knowledge, enabling the LLM agent to perform root cause analysis and generate actionable maintenance recommendations.

Given a battery charging cycle characterized by time-series measurements $\mathbf{X} = \{V(t), I(t), T(t), SOC(t)\}$, where $V(t)$, $I(t)$, $T(t)$, and $SOC(t)$ denote voltage, current, temperature, and state-of-charge signals respectively (sampled at 10s intervals), BatteryAgent produces a comprehensive diagnostic output $\mathcal{D} = \{\hat{y}, \mathcal{R}, \mathcal{M}\}$ comprising fault classification $\hat{y} \in \{\text{Normal}, \text{Abnormal}\}$, root cause explanation $\mathcal{R}$, and maintenance advice $\mathcal{M}$.

The diagnostic pipeline is formally expressed as:
\begin{equation}
\mathcal{D} = f_{\text{LLM}}\left(\Phi_{\text{SHAP}}\left(f_{\text{GBDT}}\left(g_{\text{phys}}(\mathbf{X})\right)\right); \mathcal{K}\right)
\label{eq:pipeline}
\end{equation}
where $g_{\text{phys}}(\cdot)$ extracts physics-informed features, $f_{\text{GBDT}}(\cdot)$ performs gradient boosting classification, $\Phi_{\text{SHAP}}(\cdot)$ computes attributions, and $f_{\text{LLM}}(\cdot; \mathcal{K})$ denotes LLM reasoning conditioned on knowledge base $\mathcal{K}$.

\subsection{Physics Perception Layer}

Unlike end-to-end deep learning approaches that implicitly learn representations from raw data, this layer implements \textit{domain knowledge injection} by encoding electrochemical principles into expert-defined features. This physics-informed design ensures physically meaningful and interpretable representations while reducing input dimensionality from high-frequency time-series to a compact feature vector.

We construct mechanism-driven features based on electrochemical principles, selecting 10 representative features through domain expert evaluation and correlation analysis. The feature extractor $g_{\text{phys}}: \mathbf{X} \rightarrow \mathbf{F} \in \mathbb{R}^{10}$ maps raw sensor data to a compact representation categorized into three domains (Table~\ref{tab:features}).

\subsubsection{Usage History Features}
These features characterize battery aging and usage patterns. The \textit{cycle number} $f_{\text{cyc}}$ represents the cumulative charge-discharge cycles, directly correlating with capacity fade and increased internal resistance. The \textit{constant-current (CC) phase ratio} $f_{\text{cc}}$ quantifies the proportion of charging time spent in CC mode versus constant-voltage (CV) mode~\cite{c1}. During standard CC-CV charging, a healthy battery maintains a longer CC phase; as degradation progresses, the CV phase extends due to increased polarization:
\begin{equation}
f_{\text{cc}} = \frac{t_{\text{CC}}}{t_{\text{CC}} + t_{\text{CV}}}
\end{equation}
where $t_{\text{CC}}$ and $t_{\text{CV}}$ denote the durations of CC and CV phases, respectively. A decreasing $f_{\text{cc}}$ indicates reduced charge acceptance capability. The \textit{maximum state-of-charge} $f_{\text{soc}}$ captures charging depth; values approaching 100\% increase stress on electrode materials and accelerate degradation.

\subsubsection{Voltage Characteristics}
These features evaluate voltage performance and consistency across the battery pack. The \textit{pack-to-cell voltage ratio} $f_{\text{vr}}$ is defined as:
\begin{equation}
f_{\text{vr}} = \frac{1}{N}\sum_{k=1}^{N} \frac{V_{\text{pack}}(k)}{n \cdot V_{\max,\text{cell}}(k)}
\end{equation}
where $V_{\text{pack}}(k)$ is the total pack voltage at sample $k$, $V_{\max,\text{cell}}(k)$ is the maximum cell voltage, $n$ is the number of series-connected cells, and $N$ is the number of time samples. Values closer to unity indicate better cell voltage consistency, while smaller values imply larger dispersion across cells. The \textit{voltage correlation coefficient} $f_{\text{corr}}$ measures the average Pearson correlation between individual cell voltage sequences and the pack mean voltage; lower values indicate asynchronous voltage evolution, which may be consistent with localized faults or cell inconsistency. The \textit{initial minimum voltage} $f_{\text{v0}}$ records the lowest cell voltage at charging onset, reflecting the weakest cell's pre-charge state. The \textit{voltage slope} $f_{\beta}$ is obtained from linear regression of voltage over time, capturing the overall charging rate trend.

\subsubsection{Thermal Dynamics}
These features assess heat generation and dissipation characteristics. The \textit{maximum temperature difference} $f_{\Delta T} = \max_k(T_{\max}(k) - T_{\min}(k))$ captures spatial thermal gradients across the pack, where $T_{\max}(k)$ and $T_{\min}(k)$ denote the maximum and minimum measured temperatures at sample $k$. Large gradients indicate uneven heat distribution or localized hotspots~\cite{c3}. The \textit{maximum temperature rate} $f_{\dot{T}} = \max_k(dT/dt|_k)$ represents the peak heating rate during charging; elevated values suggest abnormal heat generation or insufficient dissipation. The \textit{terminal temperature} $f_{T_{\text{end}}}$ records the temperature at charging completion; elevated values indicate excessive heat accumulation.

\begin{table}[htbp]
\caption{Summary of Mechanism-Driven Features}
\begin{center}
\renewcommand{\arraystretch}{1.4} 
\begin{tabular}{|c|c|l|}
\hline
\textbf{Category} & \textbf{Symbol} & \textbf{Physical Interpretation} \\
\hline
\multirow{3}{*}{\textit{Usage History}} 
& $f_{\text{cyc}}$ & Cumulative charge-discharge cycles \\
\cline{2-3}
& $f_{\text{cc}}$ & CC-phase ratio; degradation indicator \\
\cline{2-3}
& $f_{\text{soc}}$ & Maximum SOC; overcharge risk \\
\hline
\multirow{4}{*}{\textit{Voltage}} 
& $f_{\text{vr}}$ & Pack-to-cell voltage ratio \\
\cline{2-3}
& $f_{\text{corr}}$ & Inter-cell voltage correlation \\
\cline{2-3}
& $f_{\text{v0}}$ & Initial minimum cell voltage \\
\cline{2-3}
& $f_{\beta}$ & Voltage slope over time \\
\hline
\multirow{3}{*}{\textit{Thermal}} 
& $f_{\Delta T}$ & Maximum temperature difference \\
\cline{2-3}
& $f_{\dot{T}}$ & Maximum temperature rate \\
\cline{2-3}
& $f_{T_{\text{end}}}$ & Terminal temperature \\
\hline
\end{tabular}
\label{tab:features}
\end{center}
\end{table}

\subsubsection{Numeric-to-Semantic Bridge}
We construct a knowledge mapping $\kappa_i: f_i \mapsto \mathcal{S}_i$ associating each feature with fault mechanisms based on domain expertise (Table~\ref{tab:correlation}), enabling the LLM to trace SHAP attributions to physical root causes.

\begin{table}[htbp]
\centering
\caption{Feature-Fault Correlation Matrix}
\begin{threeparttable}
    \setlength{\tabcolsep}{8.7pt}
    \renewcommand{\arraystretch}{1.35} 
    \begin{tabular}{|c|c|c|c|c|c|c|} 
    \hline
    \textbf{Feature} & \textbf{ISC} & \textbf{TR} & \textbf{CF} & \textbf{CD} & \textbf{TM} & \textbf{BMS} \\
    \hline
    $f_{\text{cyc}}$ & $\circ$ & $\circ$ & $\bullet$ & $\bullet$ & $\circ$ & $\circ$ \\
    \hline
    $f_{\text{cc}}$ & $\circ$ & $\circ$ & $\bullet$ & $\circ$ & $\circ$ & $\bullet$ \\
    \hline
    $f_{\text{soc}}$ & $\circ$ & $\bullet$ & $\bullet$ & $\circ$ & $\circ$ & $\bullet$ \\
    \hline
    $f_{\text{vr}}$ & $\bullet$ & $\circ$ & $\circ$ & $\bullet$ & $\circ$ & $\bullet$ \\
    \hline
    $f_{\text{corr}}$ & $\bullet$ & $\circ$ & $\bullet$ & $\bullet$ & $\circ$ & $\circ$ \\
    \hline
    $f_{\text{v0}}$ & $\bullet$ & $\circ$ & $\bullet$ & $\bullet$ & $\circ$ & $\circ$ \\
    \hline
    $f_{\beta}$ & $\circ$ & $\circ$ & $\bullet$ & $\circ$ & $\circ$ & $\bullet$ \\
    \hline
    $f_{\Delta T}$ & $\bullet$ & $\bullet$ & $\circ$ & $\circ$ & $\bullet$ & $\circ$ \\
    \hline
    $f_{\dot{T}}$ & $\circ$ & $\bullet$ & $\circ$ & $\circ$ & $\bullet$ & $\circ$ \\
    \hline
    $f_{T_{\text{end}}}$ & $\circ$ & $\bullet$ & $\circ$ & $\circ$ & $\bullet$ & $\circ$ \\
    \hline
    \end{tabular}
    \begin{tablenotes}
        \footnotesize
        \item ISC: Internal Short Circuit; TR: Thermal Runaway; CF: Capacity Fade; CD: Consistency Degradation; TM: Thermal Management; BMS: BMS Fault.
        \item $\bullet$: strong correlation; $\circ$: weak correlation. Matrix constructed based on domain expertise and prior mechanism studies.
    \end{tablenotes}
\end{threeparttable}
\label{tab:correlation}
\end{table}

\subsection{Detection \& Attribution Layer}

Given feature matrix $\mathbf{F} \in \mathbb{R}^{10}$, we employ a GBDT classifier to obtain fault predictions. To enable interpretable diagnostics, SHAP is applied to quantify feature contributions. For each feature $f_j$, the SHAP value $\phi_j$ satisfies the local accuracy property $f(\mathbf{F}) = \phi_0 + \sum_j \phi_j$, ensuring attribution completeness. Features are ranked by normalized contributions $w_j = |\phi_j|/\sum_k|\phi_k|$, and the top-$k$ contributors are selected for semantic interpretation in the subsequent reasoning layer.

\subsection{Reasoning \& Diagnosis Layer}

\subsubsection{LLM Agent Architecture}
The LLM agent (DeepSeek-R1) operates on a structured prompt:
\begin{equation}
\mathcal{P} = [\text{Rules}] \oplus [\text{SHAP}] \oplus [\text{Template}]
\end{equation}
where [Rules] contains mechanism-driven knowledge base $\mathcal{K} = \{(f_i, \kappa(f_i))\}_{i=1}^{10}$ with physical interpretations, [SHAP] provides the GBDT prediction $\hat{y}$ and top-$k$ feature contributions $\{(f_j, \phi_j, w_j)\}$, and [Template] specifies the output format $\mathcal{D} = \{\mathcal{D}_{\text{result}}, \mathcal{D}_{\text{cause}}, \mathcal{D}_{\text{advice}}\}$.

The semantic bridge ensures \textit{dual anchoring}: (1) \textit{Quantitative Anchoring} via SHAP prevents hallucination by grounding explanations in actual model behavior; (2) \textit{Semantic Anchoring} via $\mathcal{K}$ enables physically meaningful root cause analysis.

\subsubsection{Probability Calibration via Token Likelihood}
When probability estimation is required, we leverage the LLM's token generation mechanism to calibrate GBDT predictions. Specifically, we compute the likelihood ratio of diagnostic tokens:
\begin{equation}
P_{\text{LLM}}(\text{Abnormal}|\mathcal{P}) = \frac{P(t_{\text{abnormal}}|\mathcal{P})}{P(t_{\text{abnormal}}|\mathcal{P}) + P(t_{\text{normal}}|\mathcal{P})}
\end{equation}
where $P(t_{\text{abnormal}}|\mathcal{P})$ and $P(t_{\text{normal}}|\mathcal{P})$ denote the generation probabilities of ``Abnormal'' and ``Normal'' tokens conditioned on prompt $\mathcal{P}$. This calibrated probability integrates both quantitative SHAP evidence and semantic reasoning, providing more reliable confidence estimates for boundary cases where GBDT exhibits uncertainty.

For standard diagnostic tasks without explicit probability requirements, the LLM directly generates the diagnosis report based on the structured prompt, bypassing probability computation for efficiency.


\section{Experimental Setup}

\subsection{Dataset}

We conduct experiments on the large-scale real-world EV battery dataset released by Zhang~et al.~\cite{b11}. This publicly available benchmark was specifically constructed to facilitate the development and validation of fault detection algorithms under realistic operating conditions. 

The dataset comprises over 690,000 charging segments collected from 347 distinct vehicles, including 292 vehicles labeled as ``Normal'' and 55 vehicles with confirmed battery faults labeled as ``Abnormal.'' 
Each charging segment contains time-series measurements from the Battery Management System (BMS), including voltage, current, and temperature sampled at 10-second intervals. This established benchmark enables direct comparison with state-of-the-art methods evaluated on the same dataset.

\subsection{Experimental Design}

\subsubsection{Data Partitioning Strategy}
Data partitioning is performed at the \textit{vehicle level}, where all charging segments from a given vehicle are exclusively assigned to either the training or validation set. This strategy strictly prevents data leakage and ensures that evaluation metrics reflect generalization capability to previously unseen vehicles rather than memorization of vehicle-specific patterns.

\subsubsection{Model Configuration}
We employ LightGBM as the GBDT implementation with default hyperparameters. For LLM-based refinement, samples with prediction confidence below 0.05 are selected for probability calibration, and the top-$k$ SHAP contributors ($k=8$) are extracted to construct the reasoning prompt, balancing diagnostic informativeness with prompt conciseness.

\subsection{Evaluation Metrics}

\subsubsection{Area Under ROC Curve (AUROC)}
AUROC quantifies the model's ability to discriminate between normal and abnormal samples across all classification thresholds, providing a comprehensive and threshold-independent measure of detection performance.

\subsubsection{Average Direct Cost}
Following the cost-sensitive evaluation framework proposed by Zhang~et al.~\cite{b11}, we compute the expected operational cost as:
\begin{equation}
C = p(1-q_{\text{TP}})c_f + [pq_{\text{TP}} + (1-p)q_{\text{FP}}]c_r
\end{equation}
where $q_{\text{TP}}$ and $q_{\text{FP}}$ denote the true positive rate and false positive rate, respectively; $p=0.038\%$ represents the empirical fault prevalence in the dataset; $c_f=5$ million CNY is the estimated cost of an undetected fault (including vehicle damage, potential injuries, and liability); and $c_r=8$ thousand CNY is the inspection cost per flagged vehicle. This metric captures the practical trade-off between missed detections and false alarms in real-world deployment scenarios.

\section{Results and Analysis}

This section presents a comprehensive evaluation of BatteryAgent through three complementary experiments: (1) quantitative comparison with state-of-the-art methods, (2) ablation study to analyze individual component contributions, and (3) case study demonstrating fine-grained diagnostic capabilities.

\subsection{Comparison with State-of-the-Art Methods}

Table~\ref{tab:sota} summarizes the performance comparison between BatteryAgent and representative baseline methods spanning four categories: specialized deep learning models for battery diagnostics (BatteryBERT~\cite{d0}, DyAD~\cite{b11}), general-purpose anomaly detection methods (AE, GDN, SVDD), time-series foundation models (TimesFM~\cite{d1}, Chronos~\cite{d2}, Time-LLM~\cite{d3}), and traditional statistical approaches (Gaussian Process (GP), Variation Evaluation (VE)).

\begin{table}[htbp]
    \setlength{\tabcolsep}{8.9pt}
    \renewcommand{\arraystretch}{1.35} 
\caption{Comparison with State-of-the-Art Methods}
\begin{center}
\begin{tabular}{|l|c|c|}
\hline
\textbf{Algorithm} & \textbf{AUROC (\%)} & \textbf{Avg. Cost (CNY)} \\
\hline
\textbf{BatteryAgent} & \textbf{98.6} & \textbf{93} \\
\hline
BatteryBERT & 94.5 & 229 \\
\hline
DyAD & 88.6 & 850 \\
\hline
TimesFM & 79.4 & 1050 \\
\hline
Chronos & 78.7 & 1090 \\
\hline
AE & 72.8 & 1330 \\
\hline
Time-LLM & 72.5 & 1240 \\
\hline
GDN & 70.3 & 1260 \\
\hline
GP & 66.6 & 1620 \\
\hline
VE & 55.6 & 1690 \\
\hline
SVDD & 51.5 & 1520 \\
\hline
\end{tabular}
\label{tab:sota}
\end{center}
\end{table}

BatteryAgent achieves an AUROC of 98.6\%, representing a 4.1 percentage point improvement over the second-best method (BatteryBERT). More notably, the average direct cost is reduced to 93 CNY, corresponding to a 59.4\% reduction compared to BatteryBERT (229 CNY) and exceeding 94\% reduction relative to classical methods. Among the baselines, specialized battery diagnostic models (BatteryBERT, DyAD) outperform general-purpose methods, confirming the importance of domain-specific design. However, time-series foundation models (TimesFM, Chronos, Time-LLM), despite their strong performance on general temporal tasks, achieve only moderate AUROC on battery fault detection, indicating that domain-agnostic representations are insufficient for this task. These results validate the effectiveness of integrating domain knowledge injection with LLM-based reasoning: the physics-informed features provide discriminative representations, while the LLM reasoning layer refines predictions on ambiguous boundary cases.

\subsection{Ablation Study}

\begin{figure}[htbp]
\centerline{\includegraphics[width=\columnwidth]{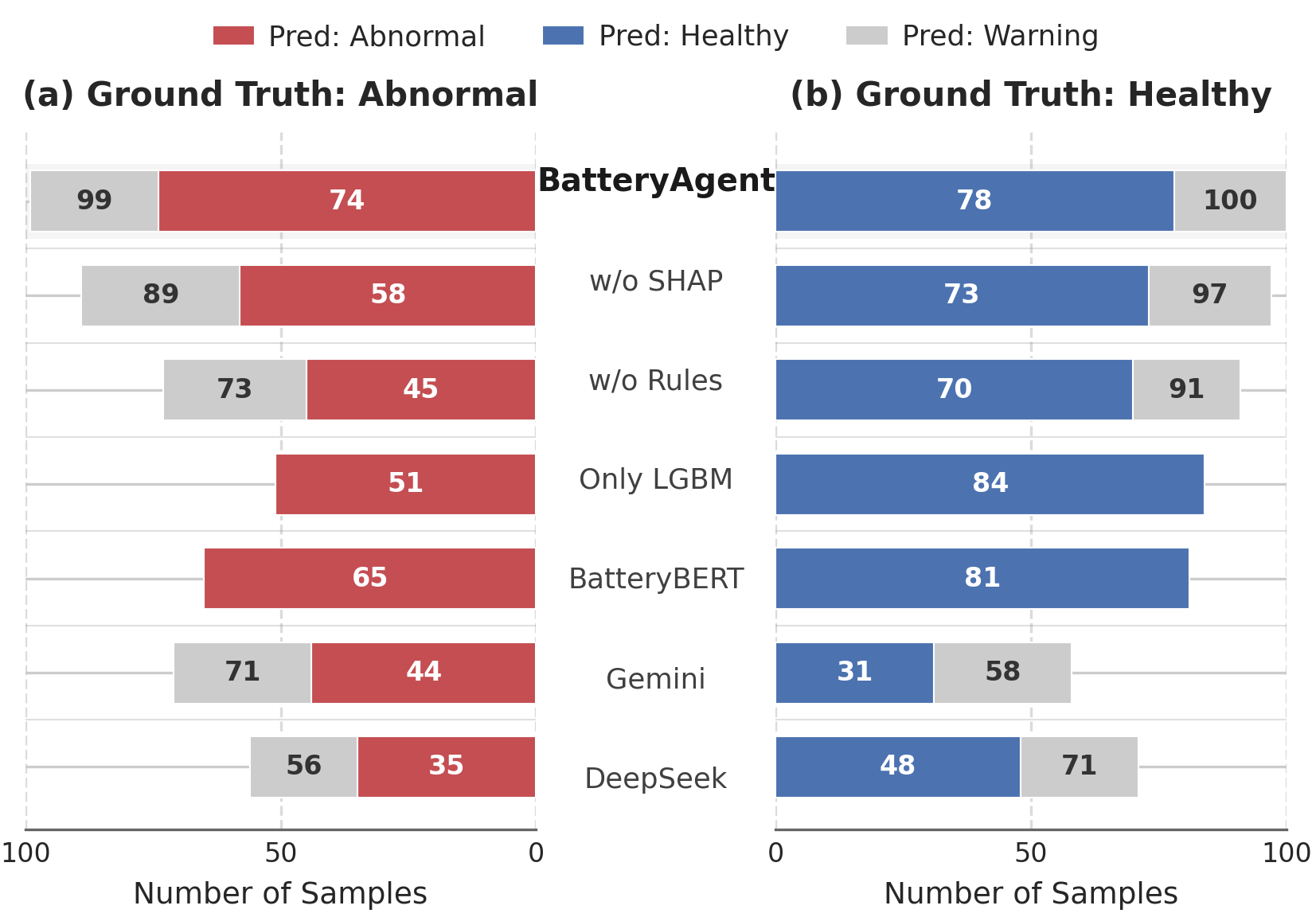}}
\caption{Ablation study results showing classification distributions. (a) Ground truth: Abnormal samples ($N=100$). (b) Ground truth: Healthy samples ($N=100$). The ``warning'' category represents cases where the model expresses uncertainty.}
\label{fig:ablation}
\end{figure}

To quantify the contribution of each architectural component, we conduct systematic ablation experiments on a balanced subset comprising 100 abnormal and 100 normal samples. It is worth noting that the original dataset employs coarse-grained binary labels, where all cycles of a faulty vehicle are marked as ``abnormal'' regardless of the degradation stage. Consequently, the ``warning'' category inferred by the LLM is considered a valid intermediate classification. It captures the transitional phase of fault evolution—where early anomalies exist but have not reached critical thresholds—effectively addressing the temporal granularity mismatch in the ground truth labels.
Fig.~\ref{fig:ablation} illustrates the classification distribution under different configurations.

The ablation results reveal several key findings. First, removing SHAP attribution increases false negatives from 1 to 11 on abnormal samples and introduces 3 false positives on normal samples, demonstrating that quantitative attribution anchoring is essential for grounding LLM reasoning in empirical evidence. Second, excluding the mechanism-driven knowledge base results in a substantial increase in false negatives (from 1 to 27), confirming that domain knowledge injection plays a critical role in accurate fault identification. Third, the complete BatteryAgent framework achieves near-optimal performance with only 1 false negative and 0 false positives, whereas the LGBM-only baseline exhibits 49 false negatives and 16 false positives, underscoring the synergistic benefit of the hierarchical architecture. Notably, BatteryBERT, despite being a domain-specific deep learning model, yields 35 false negatives and 19 false positives, indicating that end-to-end learning without explicit physical grounding remains inferior to our hybrid approach. Finally, general-purpose LLMs without domain-specific grounding exhibit significantly degraded performance (Gemini: 29\% false negative rate, 42\% false positive rate; DeepSeek: 44\% false negative rate, 30\% false positive rate), highlighting the necessity of the numeric-to-semantic bridge for reliable diagnostics.

\subsection{Fine-Grained Diagnostic Capability}

Beyond binary classification, we evaluate BatteryAgent's ability to provide multi-dimensional fault characterization, which is essential for enabling targeted maintenance rather than generic alerts. The six diagnostic dimensions correspond to the fault types defined in the knowledge base (Table~\ref{tab:correlation}). We analyze 50 consecutive charging segments from vehicle ID 405 (labeled ``Abnormal'' in the ground truth) and examine the consistency of predicted fault severity across these dimensions.

\begin{figure}[htbp]
\centerline{\includegraphics[width=0.85\columnwidth]{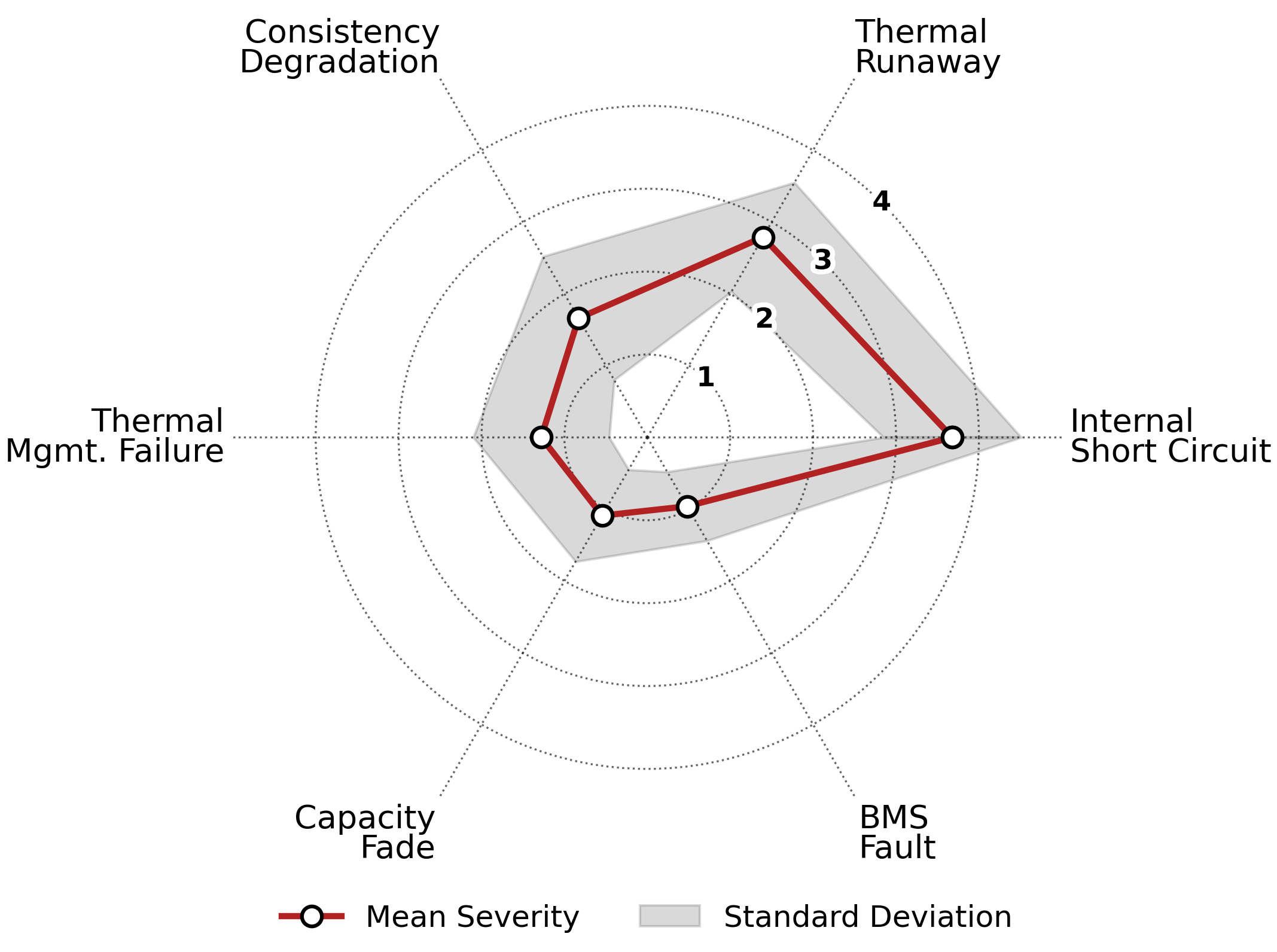}}
\caption{Fault severity profile of vehicle 405 over 50 charging segments. Solid line: mean rating; shaded region: standard deviation. Higher values indicate greater fault severity (scale: 0--5).}
\label{fig:radar}
\end{figure}

As shown in Fig.~\ref{fig:radar}, BatteryAgent consistently identifies Internal Short Circuit (mean severity = 3.68, $\sigma$ = 0.72) and Thermal Runaway risk (mean = 2.79, $\sigma$ = 0.86) as the dominant fault indicators, while Capacity Fade (mean = 1.09) and BMS Fault (mean = 0.96) remain at low severity levels. This pattern aligns with electrochemical principles: internal short circuits often elevate local heat generation, thereby increasing thermal runaway risk as a correlated secondary effect. The small standard deviations across all dimensions (ranging from 0.48 to 0.86) demonstrate high diagnostic consistency across multiple charging cycles. Beyond severity assessment, the LLM agent traces anomalies to physical root causes via SHAP attributions and provides actionable maintenance recommendations. This capability enables targeted maintenance prioritization and represents a significant advancement from binary anomaly detection toward comprehensive battery health profiling---achieved without requiring fine-grained fault annotations in the training data.

\section{Conclusion}

This paper presents BatteryAgent, a hierarchical diagnostic framework that addresses the fundamental limitations of existing battery fault detection methods by synergizing physics-informed feature engineering with large language model reasoning. The proposed framework introduces three key innovations: (1) a domain knowledge injection mechanism that encodes electrochemical principles into interpretable features, (2) a SHAP-based attribution layer that provides quantitative grounding for downstream reasoning, and (3) a numeric-to-semantic bridge that enables LLMs to perform physically meaningful root cause analysis.

Extensive experiments on a large-scale real-world EV battery dataset demonstrate that BatteryAgent achieves state-of-the-art detection performance with an AUROC of 98.6\%, while reducing average operational costs by 59.4\% compared to the best baseline. The ablation study confirms the synergistic contribution of each component, and the case study validates the framework's capability to provide consistent, multi-dimensional fault characterization without requiring fine-grained annotations.

The proposed approach represents a methodological advancement from reactive binary fault detection toward proactive, interpretable diagnostics. Future work will explore the integration of additional sensor modalities and the extension to predictive maintenance applications in industrial battery energy storage systems.

\end{document}